\documentclass[conference]{IEEEtran}
\IEEEoverridecommandlockouts
\usepackage{cite}
\usepackage{amsmath,amssymb,amsfonts}
\usepackage{algorithmic}
\usepackage{graphicx}
\usepackage{textcomp}
\def\BibTeX{{\rm B\kern-.05em{\sc i\kern-.025em b}\kern-.08em
    T\kern-.1667em\lower.7ex\hbox{E}\kern-.125emX}}
\renewcommand\IEEEkeywordsname{Keywords}
\begin{document}

\title{Recent Advances in SQL Query Generation: A Survey}

\author{
\IEEEauthorblockN{Jovan Kalajdjieski}
\IEEEauthorblockA{
\textit{Faculty of Computer Science} \\ 
\textit{and Engineering} \\ 
\textit{Ss. Cyril and Methodius University}\\
Skopje, North Macedonia \\
jovan.kalajdzhieski@finki.ukim.mk\vspace{-15pt}}
\and
\IEEEauthorblockN{Martina Toshevska}
\IEEEauthorblockA{
\textit{Faculty of Computer Science} \\ 
\textit{and Engineering}\\ 
\textit{Ss. Cyril and Methodius University}\\
Skopje, North Macedonia \\
martina.toshevska@finki.ukim.mk\vspace{-15pt}}
\and
\IEEEauthorblockN{Frosina Stojanovska}
\IEEEauthorblockA{
\textit{Faculty of Computer Science} \\ 
\textit{and Engineering}\\ 
\textit{Ss. Cyril and Methodius University}\\
Skopje, North Macedonia \\
frosina.stojanovska@finki.ukim.mk\vspace{-15pt}}
}

\maketitle

\begin{abstract}
Natural language is hypothetically the best user interface for many domains. However, general models that provide an interface between natural language and any other domain still do not exist. Providing natural language interface to relational databases could possibly attract a vast majority of users that are or are not proficient with query languages. With the rise of deep learning techniques, there is extensive ongoing research in designing a suitable natural language interface to relational databases.

This survey aims to overview some of the latest methods and models proposed in the area of SQL query generation from natural language. We describe models with various architectures such as convolutional neural networks, recurrent neural networks, pointer networks, reinforcement learning, etc. Several datasets intended to address the problem of SQL query generation are interpreted and briefly overviewed. In the end, evaluation metrics utilized in the field are presented mainly as a combination of execution accuracy and logical form accuracy.
\end{abstract}


\vspace{-9pt}

\section{Introduction}

\vspace{-4.5pt}

The possibility to use a natural language statement to query a database has the potential to attract a vast majority of users that are not proficient in using query languages such as the Structured Query Language (SQL). This language is the main query language for relational databases currently in use. The problem of text to SQL mapping could be viewed as a Semantic Parsing problem \cite{andreas2013semantic}, which is defined as transforming a natural language input into a machine-interpretable representation. Semantic parsing is a long-standing question and is a well-studied problem in Natural Language Processing (NLP). As such, it has attracted much attention both from academia and from the industry, especially translating natural language into SQL queries. A large amount of the data in today's age is stored in relational databases for applications ranging from financial and e-commerce domains to medical domains. Therefore, it comes as no surprise that querying a database using natural language has many different applications. It also opens up the prospects of having self-serving dashboards and dynamic analytics, where people not accustomed to the SQL language could use it to get the most relevant information for their business. The task of translating natural language to SQL has many related tasks such as code generation and schema generation. All these tasks could be ultimately combined to form a general task of translating natural language to a complete application.

There have been a variety of methods proposed to tackle the semantic parsing problem such as: rule-based \cite{popescu2003towards}, unsupervised \cite{goldwasser2011confidence}, supervised \cite{zelle1995using}, response based\cite{liang2011learning} and many others. However, the problem of generating SQL is more challenging than the traditional semantic parsing problem. A short natural language question could require joining multiple tables or having multiple filtering conditions. This requires more context based approaches.

For that purpose, in recent years, with the extensive development of deep learning techniques, especially convolutional and recurrent neural networks, the results are drastically improving. There have been quite a few researches attempting to generate data processing results by directly linking records in the tables to the semantic meaning of the natural language input, such as \cite{neelakantan2016learning} and \cite{yin2016neural}. However, these attempts are not scalable to big tables and are not reusable when the database schema is changed. More recent approaches use only the natural language input and the database schema and metadata to generate the queries. We review the most recent approaches in our research. Furthermore, the release of large annotated datasets containing questions and the corresponding database queries has additionally enhanced the ability to use deep learning or supervised techniques to tackle this problem. This has enabled the problem to evolve into a more complex task where the approaches should be domain independent and involving multiple tables with complex queries.

In this paper, we provide an extensive research on the most used datasets, as well as the most recent approaches applied on these datasets to handle the problem of generating SQL queries from natural language input. The main motivation of this paper is to provide a comprehensive explanation and analysis of the most recent methods to handle the task of generating SQL using natural language as well as the different datasets and evaluation techniques used. The rest of the paper is organized as follows. Section~\ref{sec:datasets} describes the different datasets that have been used in the approaches described. A comprehensive explanation of the different methods for generating SQL queries from text is presented in Section~\ref{sec:methods}. The different evaluation methods used for this problem are outlined in Section~\ref{sec:evaluation}. Finally, Section~\ref{sec:conclusion} concludes the paper.

\vspace{-4pt}

\section{Text-to-SQL Datasets}
\label{sec:datasets}

\vspace{-2pt}

The datasets designed for semantic parsing of natural language sentences to SQL queries are composed of annotated complex questions and SQL queries. The sentences are questions for a specific domain, and the answers for these questions are derived from existing databases. Therefore, the particular question is connected with an SQL query. The execution of the SQL query extracts the answer from the existing database/s.

Nowadays, there are several semantic parsing datasets developed for SQL query mapping. All of the different datasets vary in several aspects. Table~\ref{tab1:datasets} provides detailed statistics of the most used datasets among researchers. The early developed datasets concentrate on one domain and one database: ATIS~\cite{data-atis-original}, GeoQuery~\cite{data-geo-query}, Restaurants~\cite{data-restaurants-logic, data-atis-original}, Academic~\cite{data-academic}, Scholar~\cite{data-atis-geography-scholar}, Yelp~\cite{data-sql-imdb-yelp}, IMDB~\cite{data-sql-imdb-yelp} and Advising\footnote{https://github.com/jkkummerfeld/text2sql-data, last visited: 05.05.2020}~\cite{data-sql-advising}. 

The newest datasets, WikiSQL\footnote{https://github.com/salesforce/WikiSQL, last visited: 05.05.2020}~\cite{zhongSeq2SQL2017} and Spider\footnote{https://github.com/taoyds/spider, last visited: 05.05.2020}~\cite{yu2018spider}, are cross-domain context-independent with a larger size. Also, newer datasets have a greater number of questions and more comprehensive queries. The size of the datasets is crucial for the proper model evaluation. Unseen complex queries in the test sets can evaluate the model generalization ability. Authors in~\cite{data-sql-advising} show that the generalisability of the systems is overstated by the traditional data splits. The WikiSQL dataset contains a large number of questions and SQL queries, yet these SQL queries are simple and concentrated on single tables~\cite{yu2018spider}. The Spider dataset contains a more modest number of questions and SQL queries than WikiSQL. However, these questions are more complex, and the SQL queries include different SQL clauses such as join of tables and nested query~\cite{yu2018spider}.

The SParC~\cite{yu2019sparc} and CoSQL~\cite{yu2019cosql} are the extension of the Spider dataset that are created for contextual cross-domain semantical parsing and conversational dialog text-to-SQL system. These new aspects open new and significant challenges for future research in this domain.

\vspace{-4pt}

\begin{table*}[htbp]
\caption{Text-to-SQL Datasets}
\begin{center}
\begin{tabular}{|l|c|c|c|c|c|c|}
\hline
\textbf{Dataset} & \textbf{Year} & \textbf{Domain(s)} & \textbf{Databases} & \textbf{Tables} & \textbf{Questions} & \textbf{Queries} \\
\hline
ATIS~\cite{data-atis-original} & 1994 & air travel information & 1 & 32 & 5,280 & 947 \\
\hline
GeoQuery~\cite{data-geo-query} & 2001 & US geography & 1 & 8 & 877 & 247 \\
\hline
Restaurants~\cite{data-restaurants-logic, data-restaurants-original} & 2003 & restaurants, food types, locations & 1 & 3 & 378 & 378 \\
\hline
Academic~\cite{data-academic} & 2014 & Microsoft Academic Search (MAS) & 1 & 15 & 196 & 185 \\
\hline
Scholar~\cite{data-atis-geography-scholar} & 2017 & academic publications & 1 & 7 & 817 & 193 \\
\hline
Yelp~\cite{data-sql-imdb-yelp} & 2017 & Yelp website & 1 & 7 & 128 &  110 \\
\hline
IMDB~\cite{data-sql-imdb-yelp} & 2017 & Internet
Movie Database & 1 & 16 & 131 & 89 \\
\hline
WikiSQL~\cite{zhongSeq2SQL2017} & 2017 & / & 26,521 & 24,241 & 80,654 & 77,840 \\
\hline
Advising~\cite{data-sql-advising} & 2018 & student course information & 1 & 10 & 4,570 & 211 \\
\hline
Spider~\cite{yu2018spider} & 2018 & 138 different domains & 200 & 645 & 10,181 & 5,693 \\
\hline
\end{tabular}
\label{tab1:datasets}
\end{center}
\vspace{-20pt}
\end{table*}

\section{Methods}
\label{sec:methods}

\vspace{-2pt}

With the rise of deep learning techniques, there is extensive ongoing research in designing a suitable natural language interface to relational databases. Mostly, the models in this area rely upon the encoder-decoder framework that is widely used in the field of natural language processing. The following subsections present some of the models utilized in the field. Some of the models described in this paper are publicly available which enables other researchers to evaluate or build other models upon them.

\vspace{-2pt}

\subsection{SQLNet}

\vspace{-4pt}

The order of two constraints in the WHERE clause of an SQL query does not matter, but syntactically, two queries with a different order of constraints are considered as different queries. This can affect the performance of a sequence-to-sequence model. That is what SQLNet\footnote{https://github.com/xiaojunxu/SQLNet, last visited: 05.05.2020}~\cite{xu2017sqlnet} attempts to overcome. SQLNet is a novel approach for generating SQL queries from a natural language using a sketch based approach on the WikiSQL task. The sketch is generically designed to express all the SQL queries of interest. The sketch separates the query into two different token types: keywords and slots to be filled. The slots belong to either the SELECT clause or to the WHERE clause. 

The WHERE clause is the most complex structure to predict and consists of three types of slots: column, op and value. All of these types can appear multiple times, as in real queries where we can have multiple filter conditions. When predicting the WHERE clause, the authors firstly need to predict which columns to include in the conditions. For that purpose, they generate the probability of a column name \(col\) appearing in the natural language query \(Q\) which is computed as \(P_{wherecol}(col|Q) = \sigma(u_{c}^TE_{col} + u_{q}^TE_{Q|col})\) where \(\sigma\) is the sigmoid function, \(E_{col}\) and \(E_{Q|col}\) are the embeddings of the column name and the natural language question respectively, and \(u_{c}\) and \(u_{q}\) are two column vectors of trainable variables.

The embeddings \(E_{col}\) and \(E_{Q|col}\) are computed as hidden states of a bidirectional LSTM (Long-short term memory introduced in~\cite{hochreiter1997long}) which do not share their weights which enables the decision whether to include a particular column to be independent of other columns. \(E_{Q|col}\) has an additional column attention mechanism to be able to remember the particular information useful in predicting a particular column name. After \(P_{wherecol}(col|Q)\) is computed, the next step is predicting which columns need to be included in the WHERE clause. To be more precise, the authors use a network to predict the number of column slots \(K\) by translating it into a \(N + 1\) classification problem where \(N\) is an upper bound of the number of columns to choose. 

After selecting the top-\(K\) columns in the where clause, a prediction is done to predict one of the three possible operands \{$=$, $>$, $<$\}. This prediction again uses the column attention embedding \(E_{Q|col}\), which clearly shows the logical connection between the operand and the particular column upon which the operand will be used. For every column, the method also needs to predict the value slot. Unlike the column slots, the order of the tokens matters in the value slot, so a sequence-to-sequence structure is used to predict a substring from the natural language question. The encoder phase still employs a bidirectional LSTM, while the decoder phase computes the distribution of the next token using a pointer network~\cite{vinyals2015pointer, yang2017reference} with the column attention mechanism. 

On the other side, the SELECT clause consists of two types of slots: an aggregator and a column name. Another difference is that there is only one column name in the SELECT clause, unlike multiple column name slots in the WHERE clause. However, the prediction is the same as the one done in the WHERE clause, keeping in mind that the model is only trying to predict one column. After predicting the column, the probability of an aggregator is also predicted, which shares a similar structure as the prediction done for the operation slot in the WHERE clause, since there are five possible aggregators to choose from.

\vspace{-2pt}

\subsection{Bidirectional Attention}

\vspace{-4pt}

The Bidirectional Attention model\footnote{https://github.com/guotong1988/NL2SQL, last visited: 05.05.2020}~\cite{huilin2019bidirectional}, much like SQLNet employs the sketch based approach for generating an SQL query. The model consists of four separate modules: character-level and word-level embedding module, the COLUMN-SELECT module, the AGGREGATOR-SELECT module and the WHERE module.

The character embeddings in the first module are initialized using the pre-trained character-level GloVe model with 300 dimensions and then leverage convolutional neural networks with three kernels to get the next representation of the embedding. The word embeddings are initialized using the pre-trained word-level GloVe model with size 300. The words not present in the GloVe model are initialized to 0 and not to a random value because the authors have inferred that using a random value and making it trainable makes the results decrease. Because a column may contain several words, the words of one column are encoded after applying an LSTM network.

In the COLUMN-select module, each token of the questions and the column names is represented as a one-hot vector, and then is an input to a bidirectional LSTM. Using this approach, the attention information of questions and column names is captured and then used to make a prediction over the column names.

On the other hand, the authors infer that there are five types of aggregation keywords in SQL: \textit{MAX}, \textit{MIN}, \textit{COUNT}, \textit{SUM}, \textit{AVG}. They also conclude that the column name does not impact the prediction result, so the AGGREGATION-SELECT module only needs to predict the type of aggregation using the question as input. This would translate the problem to a text classification problem, where the input text is the encoded question.

The last and most challenging part is the WHERE module. Because the order of conditions does not matter, this model employs a very similar approach like SQLNet. Firstly, the number of conditions \(K\) is predicted. The prediction once again can be viewed as a (N+1) classification problem. After the number of columns is predicted, taking questions and column names as input and leveraging the bi-attention information from the inputs, it predicts the column slots, which is the same computation like in the COLUMN-SELECT module with the only difference that in this part the top-K columns are selected for the column slots. For each column slot predicted, the model then needs to select the operator from the set of three possible operators. Again, it uses the bi-attention info from the question and the column names, but now with the addition of the prediction for the column slot. The last part is the value part where leveraging the predicted columns info, a sequence-to-sequence structure is used to generate the values by taking the predicted columns info as input.

\vspace{-2pt}

\subsection{Encoder-Decoder Framework}

\vspace{-4pt}

The grammatical structure of a language can be described using Backus Normal Form (BNF), which is a set of derivation rules, consisting of a group of symbols and expressions. The BNF specification consists of two types of symbols: terminal and non-terminal symbols. Non-terminal symbols can be substituted by a sequence of expressions. There can be more than one sequence for a non terminal symbol, divided by a vertical bar meaning that one of them needs to be selected. On the other side, as the name suggests, the terminal symbols are not substituted. The terminal symbols are usually SQL keywords, operators or a concrete value expression. The encoder-decoder framework~\cite{cai2018encoder} leverages the BNF for the purpose of translating natural language inputs to SQL queries. As the name states, it consists of two phases: encoder phase and decoder phase. 

This approach firstly starts with the encoder phase with an objective of digesting the natural language input and putting the most important information in the memory before proceeding to the next phase. For this purpose, the authors propose extracting additional semantic features that link the original words to the semantics of the SQL language. The semantic features are in fact group of labels, where each label corresponds to a terminal symbol in the BNF. In the BNF of SQL-92\footnote{https://en.wikipedia.org/wiki/SQL-92, last visited: 05.05.2020} there are four terminal symbols: derived column, table reference, value expression and string expression. The authors manually label a small group of samples with these four label types and employ conditional random fields (CRFs)~\cite{lafferty2001conditional} to build effective classifiers for these labels. 

The decoder phase employs two different techniques: including the embedding of grammar state in the hidden layer and the masking of word outputs. The first technique is used for state transition. The authors state that given a particular word in the output sequence, the grammar state of the word is the last expression of BNF this word fits in. To facilitate grammar state tracking, a binary vector structure is used to represent all possible states. The length of the vector is identical to the number of expressions in the BNF.

The second technique is used to filter out invalid words for outputting, based on short term and long term rule dependencies. At each step, the decoder chooses one rule from the candidate short-term dependencies, and one or more rules from the candidate long-term dependencies. These rules are used for rule matching, and once the decoder identifies a matching rule it generates a mask on the dictionary to block the output of words not allowed by the rule. The short-term dependency is updated according to the current grammar state as well as the last output word from the decoder. Long-term dependencies on the other hand, are updated based on the active symbols chosen by the SQL parser, maintained in the grammar state vector.

\vspace{-2pt}

\subsection{Seq2SQL}

\vspace{-3pt}

Seq2SQL\footnote{https://github.com/salesforce/WikiSQL, last visited: 05.05.2020}~\cite{zhongSeq2SQL2017} method consists of two parts: augmented pointer generator network and main Seq2SQL model. The augmented pointer network generates the content of the SQL query token-by-token by copying from the input sequence. The input sequence $x$ is composed of the following tokens: words in the question, column names in the database tables and SQL clauses. The network encodes $x$ with two-layer bidirectional LSTM network using the embeddings of its words. Next, a pointer network~\cite{vinyals2015pointer} is applied. The decoder is a two-layer unidirectional LSTM that generates one token at each timestep using the token generated in the previous step. It produces scalar attention score for each position of the input sequence. The token with the highest score is selected as next token. The second part, Seq2SQL, is composed of three different parts: Aggregation Operation, SELECT Column and WHERE Clause.

The first part, Aggregation Operation, classifies aggregation operation of the query, if any. First, scalar attention score is computed for each token in the input sequence. The vector of scores is then normalized in order to produce a distribution over the input tokens. It is computed with a Multilayer Perceptron (MLP) with cross-entropy loss. The second part, SELECT Column, points to a column in the input table. Each column name is first encoded with LSTM network such that the last encoded state of the LSTM is assumed to be representation of the specific column. With the same architecture, representation for the input question is calculated. MLP with cross-entropy loss is applied to compute score for each column conditioned on the input representation. The last part, WHERE Clause, generates the conditions for the query. For this part, reinforcement learning is applied to optimize the expected correctness of the execution result. Next token is sampled from the output distribution. When the complete query is generated, it is executed against the database. The reward is: (1) -2 if the generated query is not a valid SQL query, (2) -1 if the generated query is a valid SQL query but executes to an incorrect result, and (3) +1 if the generated query is a valid SQL query and executes to the correct result. The loss is the negative expected reward over possible WHERE clauses.

The overall model is trained using gradient descent to minimize the objective function that is the combination of the objective functions of its composing parts. However, this method does not incorporate complex SQL queries such joining tables and nested queries.

\vspace{-2pt}

\subsection{STAMP}

\vspace{-4pt}

Syntax- and Table-Aware seMantic Parser (STAMP)~\cite{sun2018semantic} is a model based on Pointer Networks~\cite{vinyals2015pointer}. It is composed of two separate bidirectional Gated Recurrent Unit (GRU) networks as encoder and decoder. An additional bidirectional RNN is used to encode the column names. The STAMP model is composed of three different channels, that are attentional neural networks: (1) SQL channel - predict SQL clause, (2) Column channel - predict column name and (3) Value channel - predict table cell. For SQL and Value channel, the input is the decoder hidden state and representation of the SQL clause. Column channel has an additional input that is the representation of the question. Feed-forward neural network is used as a switching gate for the channels.

Column-cell relation is incorporated into the model in order to improve the prediction of SELECT column and WHERE value. The representation of the column name is enhanced with cell information. The importance of a cell is measured with the number of cell words occurring in the question and then the final importance of the cell is normalized with softmax function. The vector representing the column is concatenated with weighted average of the cell vectors that belong to that column. An additional global variable to memorize the last predicted column name is added. When the switching gate selects the Value channel, the cell distribution is only calculated over the cells belonging to the last predicted column name.

The model is trained in different ways: with standard cross-entropy loss over the pairs of question and SQL query, and with reinforcement learning with policy gradient as in~\cite{zhongSeq2SQL2017}.

\vspace{-2pt}

\subsection{One-Shot Learning for Text-to-SQL Generation}

\vspace{-3pt}

A method for SQL query generation composed of template classification and slot filling is presented in~\cite{lee2019one}. The first phase, template classification consists of two networks: Candidate Search Network and Matching Network. The first network, Candidate Search Network, chooses $n$ most relevant templates. The network is a convolutional neural network and is trained to classify a natural language question where the classes represent SQL templates. For a given question, features from the layer before the final classification layer are extracted. Then, $n$ most similar vectors with the question vector are obtained using cosine similarity. The second network, Matching Network, predicts the SQL template. First, an encoder is used to embed the question. The encoder is convolutional neural network consisted of convolutional layers with different window sizes with max-pooling. The final representation of the question is a concatenation of each pooled feature. An attention-based classifier predicts the template label based on the feature vectors obtained with the Candidate Search Network.

The second phase, slot filling, is a Pointer Network~\cite{vinyals2015pointer} that fills the slots of the predicted SQL template. The encoder is bidirectional LSTM network, while the decoder is unidirectional LSTM network. The network determines the tokens by maximizing the log-likelihood of the predicted token for the given natural language question and list of variables in the SQL template. The decoder generates one token at each timestep using attention over its previous hidden state and the encoder states.

\vspace{-2pt}

\subsection{Relation-Aware Self-Attention for Text-to-SQL Parsers}

\vspace{-3pt}

This approach explained in~\cite{shin2019encoding}\footnote{https://github.com/rshin/seq2struct, last visited: 05.05.2020} is an improvement on the already existing methods so that it overcomes some crucial limitations such as: working in only one domain, working with one database schema, working with only one table or overlapping training and test sets. The main improvement of this approach focuses on the encoder part of the encoder-decoder framework already seen in previous approaches. To incorporate the relationships between schema elements in the encoder, the database schema is translated to a directed graph where each node represents either a table or a column and the edge represents the relationship between the elements. The label in the node represents the name of the table or column appropriately. The columns additionally have their type prepended. All the edges between the nodes are labeled as well to represent the exact relationship they represent. The relationships can be: (1) Column X Column Y relationship where X and Y belong to the same table or X is a foreign key for Y (or vice versa), (2) Column X Table T relationship (or vice versa) where X is the primary key of T or X is a column of T (but not a primary key), and (3) Table T Table R relationship where T has a foreign key column in R (or vice versa) or T and R have foreign keys in both directions.
%\begin{itemize}
%    \item Column X Column Y relationship where 
%        \begin{itemize}
%            \item X and Y belong to the same table
%            \item X is a foreign key for Y (or vice versa)
%        \end{itemize}
%    \item Column X Table T relationship (or vice versa)
%        \begin{itemize}
%            \item X is the primary key of T
%            \item X is a column of T (but not a primary key)
%        \end{itemize}
%    \item Table T Table R relationship
%        \begin{itemize}
%            \item T and R have foreign keys in both directions
%        \end{itemize}
%\end{itemize}
%\begin{itemize}
%    \item Column X Column Y relationship where X and Y belong to the same table or X is a foreign key for Y (or vice versa)
%    \item Column X Table T relationship (or vice versa) where X is the primary key of T or X is a column of T (but not a primary key)
%    \item Table T Table R relationship where T has a foreign key column in R (or vice versa) or T and R have foreign keys in both directions
%\end{itemize}

After obtaining the initial graph representation, bidirectional LSTM is applied to the labels in the nodes and the output of the initial and final timesteps of the LSTM is concatenated to obtain the embedding for the node. For the input question, bidirectional LSTM is also used. Until this point, these initial representations are independent of one another in the sense that they do not have any information which other columns or tables are present. For that purpose, a relation-aware self-attention transformation is applied to all the elements to encode the relationship between two elements. For brevity, we omit the mathematical model used to represent the relationships. 

The formulation of the relation-aware self-attention is the same as the one used in Shaw et al.~\cite{shaw2018self}. However, in this approach, it is shown that relation-aware self-attention can effectively encode more complex relationships that exist within an unordered sets of elements compared with relationships between two words. After applying this transformation, the final encoding of the columns, tables and the input question is used in the decoder. The decoder used is the same as in~\cite{yin2017syntactic}. The decoder generates the query using a depth-first traversal order in an abstract syntax tree. It outputs a sequence of production rules that expand the last generated node in the tree. The decoder does not output the FROM clause. It is recovered afterwards using hand-written rules where only the columns referred to in the remainder of the query are used. Small modifications have been made to this decoder, namely: (1) when the decoder needs to output a column a pointer network based on scaled dot-product attention is used, and (2) at each step, the decoder accesses the encoder outputs using multi-head attention.

\vspace{-4pt}

\section{Evaluation}
\label{sec:evaluation}

\vspace{-4pt}

There is no single metric for evaluation of the text-to-SQL model. One strategy is to estimate the correctness of the result for the question. This metric is called \textit{execution accuracy}~\cite{zhongSeq2SQL2017}. It compares the result from the generated SQL query and the result from the ground truth query. Then it returns the number of correct matches divided by the total number of examples in the dataset. One shortcoming of this approach is that it does not eliminate the cases when a completely different query is giving the same result as the expected, for example, the NULL result.

The second metric is the \textit{logical form accuracy}~\cite{zhongSeq2SQL2017}. This approach calculates the exact matches of the synthesized query and the ground truth query. The queries are represented as strings, and the method for comparison is the exact string match of the queries. The weakness of this approach is the penalization of the queries that are correct but do not achieve a complete string match with the ground truth query; for example, different order of the returning columns or different queries for the same purpose. To partially address this issue, the authors in~\cite{xu2017sqlnet} introduce the \textit{query match accuracy}. The predicted and ground truth queries are represented in a canonical form to perform the matching of the queries. This approach only solves the false negatives due to the ordering issue. SQL canonicalization is an approach used to eliminate the problem of the different writing style by ordering the columns and tables and using standardized aliases~\cite{data-sql-advising}.

The evaluation metric in~\cite{yu2018spider} includes \textit{component} and \textit{exact matching} of the queries. Each query is divided into components: SELECT, WHERE, GROUP BY, ORDER BY and KEYWORDS. The predicted and ground truth queries are divided and represented as subsets for each of the components, and these subsets are then compared with exact matching. However, the problem of the novel synthesized syntax for the identical logic of the SQL query is not eliminated, so the execution accuracy is needed for a comprehensive evaluation. The approach in~\cite{yu2018spider} also incorporates one novelty in the evaluation process, the difficulty of the SQL query. Dividing the results by the hardness criteria can be more informative of the general ability of the model.

None of the current metrics can be used as a standalone evaluation metric for exact evaluation and comparison of the models, so the combination of the existing metrics is essential. It is critical for the future work in this domain to incorporate the evaluation question. 

\vspace{-4pt}

\section{Conclusion}
\label{sec:conclusion}

\vspace{-4pt}

The translation of a natural language to SQL queries is a problem of semantic parsing. There are several text-to-SQL datasets developed that include natural language questions that can be answered by executing an SQL query from a database. The progression of the datasets introduces a combination of different domains with multiple databases and tables. The increase in the size of the datasets is apparent. Also, the questions are becoming more complex and in more extensive number. 

The progressions in the NLP area are reflected in the designed models of this problem. The encoder-decoder framework is incorporated to translate the natural language into an SQL query. The encoder serves for natural language processing, whereas the decoder predicts the BNF representation of the SQL output. The sketch-based approach is introduced for SQL representation for eliminating the ordering effect of sequence generation. Additional efforts incorporate attention to the bidirectional LSTM network with the sketch-based method. The augmented pointer network is also combined in the novel models. The Relation-Aware Self-Attention approach is an improvement on already existing methods to overcome several limitations. It includes a relationship graph of the database schema and self attention to encode more complex relationships.

To evaluate the models, several approaches combine the execution accuracy and logical form accuracy. The latest approaches divide the accuracy metric into component and exact matching with the additional information of the difficulty of the SQL query. 

\vspace{-6.5pt}

\bibliography{bibliography}
\bibliographystyle{ieeetr}

\end{document}